\documentclass[11pt,a4paper]{article}
\usepackage[hyperref]{acl2020}
\usepackage{times}
\usepackage{latexsym}

\usepackage{microtype}
\usepackage{graphicx}

\aclfinalcopy 

\title{Using Integrated Gradients and Constituency Parse Trees to explain Linguistic Acceptability learnt by BERT}

\author{Anmol Nayak, Hari Prasad Timmapathini\\
  ARiSE Labs at Bosch\\
  \texttt{\{Anmol.Nayak, HariPrasad.Timmapathini\}@in.bosch.com} \\}

\date{}

\begin{document}
\maketitle
\begin{abstract}
Linguistic Acceptability is the task of determining whether a sentence is grammatical or ungrammatical. It has applications in several use cases like Question-Answering, Natural Language Generation, Neural Machine Translation, where grammatical correctness is crucial. In this paper we aim to understand the decision-making process of BERT \cite{devlin-etal-2019-bert} in distinguishing between Linguistically Acceptable sentences (LA) and Linguistically Unacceptable sentences (LUA). We leverage Layer Integrated Gradients Attribution Scores (LIG) to explain the Linguistic Acceptability criteria that are learnt by BERT on the Corpus of Linguistic Acceptability (CoLA) \cite{warstadt2018neural} benchmark dataset. Our experiments on 5 categories of sentences lead to the following interesting findings: 1) LIG for LA are significantly smaller in comparison to LUA, 2) There are specific subtrees of the Constituency Parse Tree (CPT) for LA and LUA which contribute larger LIG, 3) Across the different categories of sentences we observed around 88\% to 100\% of the Correctly classified sentences had positive LIG, indicating a strong positive relationship to the prediction confidence of the model, and 4) Around 43\% of the Misclassified sentences had negative LIG, which we believe can become correctly classified sentences if the LIG are parameterized in the loss function of the model.
\end{abstract}

\section{Introduction}

Linguistic acceptability is an important criteria in Natural Language Processing and is one of the tasks in the General Language Understanding Evaluation (GLUE) benchmark \cite{wang2018glue}. With the evolution of language encoders like BERT (which leverages the multi-head self-attention mechanism \cite{vaswani2017attention} in its architecture) that have been a breakthrough in language understanding and achieved state-of-the-art results, the field of probing these architectures for understanding their behaviours has become important.

While there have been several works on interpreting and understanding the different layers of BERT with respect to lexical, syntactic and semantic behaviours \cite{jawahar2019does,lin-etal-2019-open,clark-etal-2019-bert,vashishth2019attention,rogers2020primer}, the focus on explaining the linguistic acceptability (grammaticality) learnt by BERT has been sparse. Some of the recent works have used probing tasks to understand the model's knowledge on particular grammatical features \cite{shi2016does,ettinger2016probing,tenney2019you}, relying on language model probabilities to judge grammatical acceptability on sentences that differ minimally \cite{marvin2018targeted,wilcox2019structural}, or probing the model's by training with boolean grammaticality judgement objectives \cite{linzen2016assessing,warstadt2018neural,kann2019verb,warstadt2019investigating}. These methods have made significant progress in uncovering that BERT has indeed learnt various aspects of grammatical knowledge, however their focus has not been on explaining the black box details of how BERT arrives at a grammaticality judgement. Our paper attempts to address this by explaining the model's linguistic acceptability judgement with LIG and CPT (a type of grammar tree) representations.

Attention mechanism based methods \cite{bahdanau2014neural,vaswani2017attention} provide interpretable understanding of the model's behaviour, however the attention scores cannot be solely relied upon since a feature could influence the output in multiple ways (for e.g. through memory cells, recurrent states etc. in LSTM networks). Feature attribution methods aim to understand the relationship between the model's output and the input features. They are helpful in interpreting the black-box details of neural networks and provide insights that can be used to improve model performance. While previous feature attribution methods such as DeepLift \cite{shrikumar2016not,shrikumar2017learning}, Layer-wise relevance propagation \cite{binder2016layer} and LIME \cite{ribeiro2016should} have provided interesting frameworks, they break at least one of the two axioms that are fundamental for attribution methods, namely \textit{Sensitivity} and \textit{Implementation Invariance} \cite{sundararajan2017axiomatic}.

In our paper we have chosen the Integrated Gradients (IG) \cite{sundararajan2017axiomatic} attribution method as it satisfies both the aforementioned axioms. IG is a post-hoc interpretability technique which aggregates the gradients of the input by interpolating in small steps along the straight line between a baseline (typically a vector with all zeros) and the input. A large positive or negative IG score indicates that the feature strongly increases or decreases the network output respectively, while a score close to zero indicates that the feature does not influence the network output. This can also be understood as follows: a positive score indicates that the feature tends to agree with the model’s prediction, while a negative score indicates that the feature tends to disagree with the model’s prediction. LIG are computed as the IG between the model output and a particular layer’s input or output. Our work attempts to answer the following Research Questions:

\begin{table}
\centering
\begin{tabular}{ll}
\hline
\textbf{Sentence} & \textbf{Category}\\
\hline
rebecca saw the play .&CIA\textsuperscript{LA}\\
the play saw .&CIA\textsuperscript{LUA}\\
i surprised myself .&RAA\textsuperscript{LA}\\
i surprised himself .&RAA\textsuperscript{LUA}\\
the boy is here .&SVA\textsuperscript{LA}\\
the boy are here .&SVA\textsuperscript{LUA}\\
michael read the book .&SVO\textsuperscript{LA}\\
michael the book read .&SVO\textsuperscript{LUA}\\
what did rebecca read ?&WHE\textsuperscript{LA}\\
what did rebecca read the book ?&WHE\textsuperscript{LUA}\\
\hline
\end{tabular}
\caption{Sample sentences across the 5 categories from the CoLA Targeted Test Sets.}
\label{tab:sampsent}
\end{table}

\begin{table*}
\centering
\begin{tabular}{lll}
\hline
\textbf{CPT Pattern} & \textbf{Avg. LIG} & \textbf{Category}\\
\hline
(S(NP(NN))(VP(VBD)\textbf{(NP(DT)(NN))})(.))&0.065&CIA\textsuperscript{LA}\\
\textbf{(S(NP(DT)(NN))(VP(VBD))(.))}&1.351&CIA\textsuperscript{LUA}\\
(S(NP(PRP))(VP(VBD)\textbf{(NP(PRP))})(.))&0.061&RAA\textsuperscript{LA}\\
\textbf{(S(NP(PRP))(VP(VBD)(NP(PRP)))(.))}&0.926&RAA\textsuperscript{LUA}\\
(S\textbf{(NP(DT)(NN))}(VP(VBZ)(ADVP(RB)))(.))&0.064&SVA\textsuperscript{LA}\\
(S(NP(DT)\textbf{(NN)})(VP(VBP)(ADJP(JJ)))(.))&1.067&SVA\textsuperscript{LUA}\\
(S(NP(NN))(VP(VBD)\textbf{(NP(DT)(NN))})(.))&-0.012&SVO\textsuperscript{LA}\\
(S(NP(NP(NN))(NP(DT)(NN)))(VP\textbf{(VBD)})(.))&1.226&SVO\textsuperscript{LUA}\\
(SBARQ(WHNP(WP))(SQ(VBD)(NP(NN))\textbf{(VP(VB))})(.))&0.205&WHE\textsuperscript{LA}\\
\textbf{(SBARQ(WHNP(WP))(SQ(VBD)(NP(NN))(VP(}&1.394&WHE\textsuperscript{LUA}\\
\textbf{VB)(NP(DT)(NN))))(.))}&&\\
\hline
\end{tabular}
\caption{Average normalized LIG of most frequent CPT patterns on 5 categories of Correctly classified CoLA Targeted Test Sets sentences. Subtrees in bold have the largest LIG in the respective categories.}
\label{tab:cptpatt}
\end{table*}

\begin{figure*}[h]
\centering
\includegraphics[scale = 0.465]{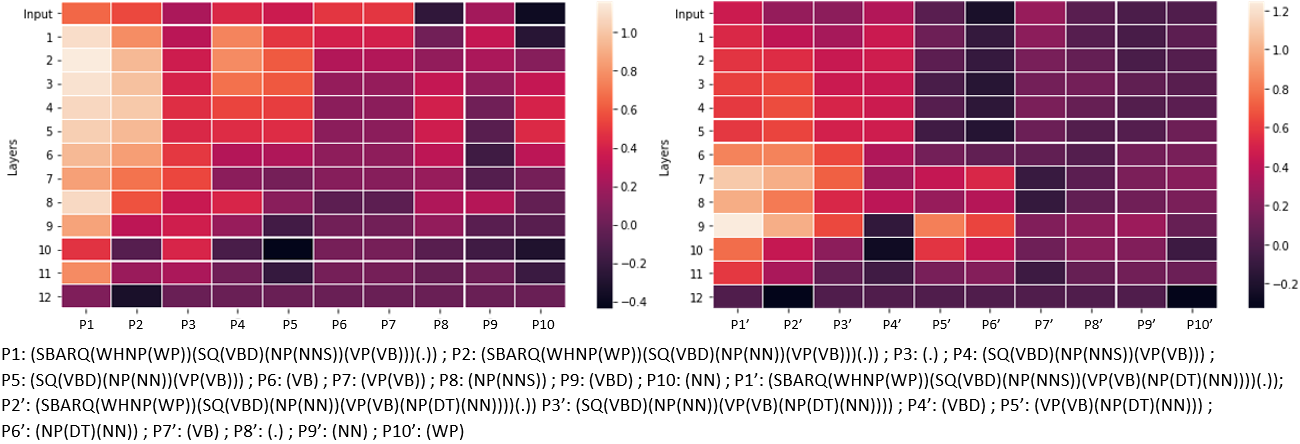}
\caption{LIG heatmaps of the top 10 scoring CPT patterns ranked in descending order based on averaged LIG across the different BERT layers for LA (Left) and LUA (Right) of the WHE category.}
\label{fig:heat}
\end{figure*}

\begin{figure}[h]
\centering
\includegraphics[scale = 0.55]{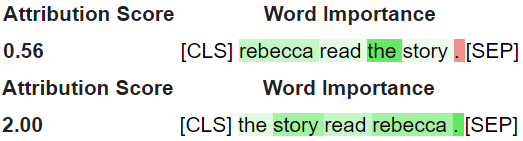}
\caption{LIG visualization for a LA sentence (Top) and LUA sentence (Bottom) of the SVO category. Green highlighted words contributed strongly towards the model output to be predicted as LA and LUA.}
\label{fig:vizligas}
\end{figure}

\begin{enumerate}
    \item Can LIG of a Constituency Parse Tree (CPT) give insights on LA vs LUA?
    \item Can LIG be reliably used to explain the Linguistic Acceptability criteria learnt by BERT?
    \item Is there a relationship between LIG and the prediction confidence of the model?
\end{enumerate}

\section{Experiment Setup}
CoLA dataset sentences have a boolean acceptability judgement, namely LA and LUA. We have used the fine-tuned BERT-Base-Uncased-CoLA model (12 encoder layers with 12 attention heads) provided by TextAttack \cite{morris2020textattack}, the Captum PyTorch Interpretability library  \cite{kokhlikyan2020captum} for computing LIG and the Stanford CoreNLP toolkit (version 4.2.1) \cite{manning2014stanford} for constructing the CPT. Integrated Gradients (IG) across the \(i^{th}\) dimension of input \(x\) and baseline \(x'\) are computed as follows: \[IG=(x_{i}-x_{i}')\times\int_{\alpha=0}^{1}\frac{\partial F(x'+\alpha\times(x-x'))}{\partial x_{i}}d\alpha\]

Captum library approximates the above integral using the Gauss-Legendre approximation algorithm over 50 uniform steps of \(\alpha \in [0,1]\). The baseline was selected as a 768 dimension zero vector. The attribution score for each word is summed across the dimensions (768 in the case of BERT-Base) and normalized using the Euclidean norm of the scores of all the words in the sentence.

We analyzed 5 different categories of sentences within the CoLA Targeted Test Sets: Causative-Inchoative Alternation (CIA), Reflexive-Antecedent Agreement (RAA), Subject-Verb Agreement (SVA), Subject-Verb-Object (SVO) and Wh-Extraction (WHE). A few sample sentences across the categories can be seen in Table~\ref{tab:sampsent}.

\begin{figure*}[h]
\centering
\includegraphics[scale = 0.49]{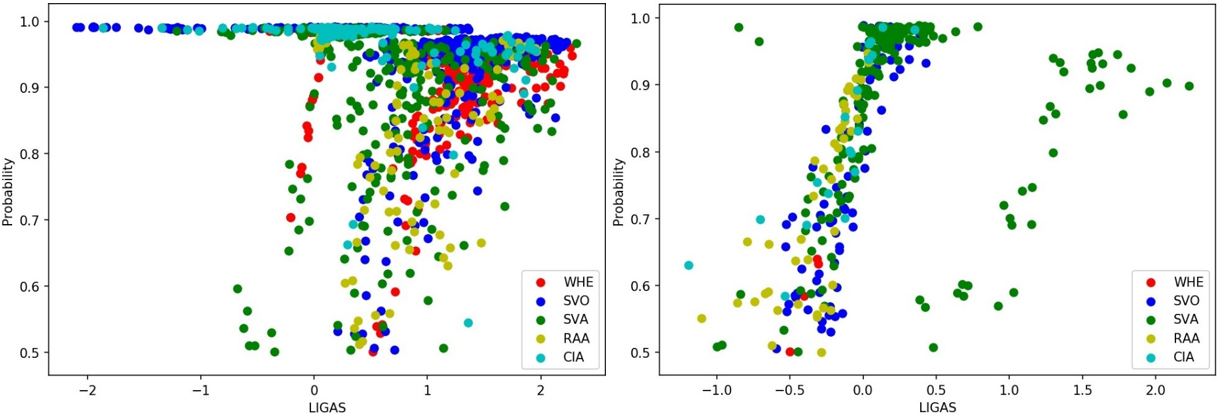}
\caption{Prediction Probability vs LIG Scatter plots for Correctly classified (Left) and Misclassified (Right) sentences.}
\label{fig:viz}
\end{figure*}

The primary focus of our experiments relied on the LIG computed between the predicted class logit and the token embedding of the words. Further we also computed LIG heatmaps with respect to the Input (Token + Segment + Position) embedding and across the 12 Encoder layer embeddings of BERT to analyze the LIG characteristics. Figure~\ref{fig:heat} shows the LIG heatmaps of the top 10 CPT patterns for the LA and LUA in the WHE category. The unique CPT patterns were extracted for the correctly classified sentences of each category, corresponding to which the LIG of each subtree were computed. LIG of a subtree is equal to the sum of the LIG of the words appearing as leaf nodes in the subtree. The results in Table~\ref{tab:cptpatt}, Table~\ref{tab:ligas}, Figure~\ref{fig:vizligas} and Figure~\ref{fig:viz} represent the LIG computed between the predicted class logit and the token embedding of the words. For Out-Of-Vocabulary words (OOV), the LIG are summed across its tokenized sub-words.


\begin{table*}
\centering
\begin{tabular}{llllllllll}
\hline
\textbf{Category} & \textbf{C} & \textbf{CC} & \textbf{MC} & \textbf{CC\textsuperscript{+}} & \textbf{CC\textsuperscript{-}} & \textbf{MC\textsuperscript{+}} & \textbf{MC\textsuperscript{-}} & \textbf{CC\textsuperscript{+}\%} & \textbf{MC\textsuperscript{+}\%}\\
\hline
CIA&182&162&20&144&18&7&13&88.88&35\\
RAA&144&100&44&100&0&2&42&100&4.54\\
SVA&676&476&200&441&35&148&52&92.64&74\\
SVO&500&400&100&362&38&54&46&90.5&54\\
WHE&520&516&4&465&51&0&4&90.11&0\\
\hline
\end{tabular}
\caption{LIG assessment for Correctly classified sentences (CC) and Misclassified (MC) sentences (C: Count, CC\textsuperscript{+}: Count of CC having positive LIG, CC\textsuperscript{-}: Count of CC having negative LIG, MC\textsuperscript{+}: Count of MC having positive LIG, MC\textsuperscript{-}: Count of MC having negative LIG, CC\textsuperscript{+}\%: Percentage of CC\textsuperscript{+} in CC, MC\textsuperscript{+}\%: Percentage of MC\textsuperscript{+} in MC).}
\label{tab:ligas}
\end{table*}

\section{LIG for Constituency Parse Tree patterns}

CPT is a type of grammar tree which captures the relations between the constituents of a sentence. We believe that analyzing the CPT patterns will give us insights into the grammatical structure of LA and LUA. Computing the LIG for CPT patterns at different subtree levels can give us an indication into the constituents which contribute largely towards making the sentence LA or LUA.

For each of the 5 category of sentences, we extracted all the CPT patterns for correctly classified sentences at every subtree level and picked the most frequent patterns at the root level (Table~\ref{tab:cptpatt}). The subtree patterns in bold are the highest ranking subtrees based on LIG. BERT has been shown to learn surface level features in the
early layers, syntactic features in the middle layers and semantic features in the higher layers \cite{jawahar2019does}. Hence, we also wanted to analyse the LIG behaviour across the 12 layers and especially the early to middle layers which are relevant for grammar understanding. Across each of the categories in the LIG heatmaps, it was seen that the top subtree CPT patterns based on token embedding LIG were also dominating across the input and encoder layers of BERT and hence were also found in the top 10 patterns. Further, it can be observed in Figure~\ref{fig:heat} that there are specific subtrees which dominate more (shades of orange) as compared to others. This characteristic is especially useful for debugging LUA as it helps us to understand which phrase contributed largely towards making it unacceptable. 

Further, it can be observed in Table~\ref{tab:cptpatt} that the LIG for LUA are significantly larger than LA. The dominating CPT subtree patterns had a large spike in the LIG for LUA in comparison to LA, indicating that linguistically acceptable patterns were not being adhered. In Figure~\ref{fig:vizligas} we can see how the different words in the LA and LUA of the SVO category contributed in varying magnitudes towards the model's prediction.

\section{LIG and Prediction confidence of the model}

We investigated to check if there is a relationship between the LIG and the prediction confidence of the model. We found that the range of correctly classified sentences having positive LIG is between 88\% to 100\% (CC\textsuperscript{+}\% in Table~\ref{tab:ligas}) indicating that whenever the input contributes strongly towards a particular class (whether it is LA or LUA), the model has a higher confidence in making the correct prediction. Around 43\% (MC\textsuperscript{-} in Table~\ref{tab:ligas}) of the total misclassified sentences had negative LIG which showed that the features disagreed with the model’s prediction. This behaviour can be observed distinctly in the Figure~\ref{fig:viz} scatter plots, where we notice that there a large number of points near the top right corner for the correctly classified sentences, and a large number of points near the bottom left corner in the case of misclassified sentences. 

We believe that this indication can be used to improve the model’s performance by parameterizing the LIG in the loss function during the later stages of the training process once the model has achieved a reasonable performance (to ensure that the gradients computed are meaningful) and hence serve as a correction mechanism for the model. This aligns with a previous work \cite{erion2021improving} which showed that axiomatic attribution priors improved model performance on many real-world tasks.

\section{Conclusion}

We have proposed a novel approach for explaining the Linguistic Acceptability criteria learnt by BERT using LIG and CPT patterns. As there is a strong relationship between LIG and the prediction confidence of the model, our future work will focus on parameterizing the LIG in the loss function and observing the model's performance.

\bibliography{anthology,acl2020}
\bibliographystyle{acl_natbib}

\end{document}